\documentclass{article}

    \PassOptionsToPackage{numbers}{natbib}

\usepackage[final]{neurips_fitml_2024}

\usepackage[utf8]{inputenc} 
\usepackage[T1]{fontenc}    
\usepackage{hyperref}       
\usepackage{url}            
\usepackage{booktabs}       
\usepackage{amsfonts}       
\usepackage{nicefrac}       
\usepackage{microtype}      
\usepackage{xcolor}         
\usepackage{graphicx}
\usepackage{array}
\usepackage{multirow}
\usepackage{algpseudocode}
\usepackage{algorithm}
\usepackage{amsmath}
\usepackage{booktabs}
\usepackage{siunitx}
\usepackage{array}
\newcolumntype{P}[1]{>{\centering\arraybackslash}p{#1}}
\newcolumntype{C}[1]{>{\centering\arraybackslash}m{#1}}
\usepackage{caption}
\title{Accelerating Direct Preference Optimization with Prefix Sharing}

\author{%
  Franklin Wang\textsuperscript{*}\\
  MIT CSAIL\\
  \texttt{fxwang@mit.edu} \\
  \And
 Sumanth Hegde\textsuperscript{*}\\
  Anyscale\\
  \texttt{sumanthrh@anyscale.com}
}

\begin{document}

\def\thefootnote{*}\footnotetext{Equal contribution.}

\maketitle

\def\thefootnote{\arabic{footnote}}

\begin{abstract}
Offline paired preference optimization algorithms have become a popular approach for fine-tuning on preference data, outperforming traditional supervised fine-tuning in various tasks. However, traditional implementations often involve redundant computations, especially for tasks with long shared prompts. We introduce prefix sharing for preference tuning, a novel technique that processes chosen and rejected responses as one sequence with a shared prefix. To prevent cross-response contamination, we use a custom block-sparse attention mask. Our method achieves $1.1$-$1.5\times$ improvement in training throughput on popular DPO datasets, without any effect on convergence. When combined with sequence packing, we observe consistent $1.3$-$1.6\times$ speedups, benefiting even datasets with smaller sequence lengths. While we focus on Direct Preference Optimization (DPO), our approach is applicable to other paired preference tuning methods. By enhancing computational efficiency, our work contributes to making preference-based fine-tuning more accessible for a wider range of applications and model sizes. We open-source our code at \url{https://github.com/frankxwang/dpo-prefix-sharing}.
\end{abstract}

\section{Introduction}

Offline paired preference optimization algorithms such as DPO \cite{rafailov2024directpreferenceoptimizationlanguage}, ORPO \cite{hong2024orpomonolithicpreferenceoptimization}, and SimPO \cite{meng2024simposimplepreferenceoptimization} have emerged as popular approaches for fine-tuning large language models (LLMs) on preference data, attaining higher performance than traditional supervised fine-tuning (SFT) in areas such as instruction-following \cite{ivison2023camelschangingclimateenhancing}, multi-step reasoning \cite{pang2024iterativereasoningpreferenceoptimization, lu2024stepcontrolleddpoleveragingstepwise, lai2024stepdpostepwisepreferenceoptimization}, agentic planning \cite{putta2024agentqadvancedreasoning}, and coding \cite{yuan2024advancingllmreasoninggeneralists, weyssow2024codeultrafeedbackllmasajudgedatasetaligning}. These algorithms leverage paired preference data, where each training sample comprises a shared prompt and two responses, with a label indicating which response is preferred. The LLM is then optimized to increase the likelihood of the chosen (preferred) response while minimizing the likelihood of the rejected (not preferred) one. This enables the LLM to learn from contrasts between the chosen and rejected samples rather than only mimicking the chosen samples through SFT.

Traditional paired preference optimization implementations batch the chosen and rejected sequences together during training. However, this results in redundant computations since each shared prompt is processed by the model twice---once for each response. This is particularly inefficient for tasks such as summarization \cite{stiennon2022learningsummarizehumanfeedback} and mathematics \cite{pal2024smaugfixingfailuremodes,yu2024metamathbootstrapmathematicalquestions} which have disproportionately long prompts compared to responses, along with multi-turn conversational datasets where the prompt consists of multiple previous interactions between the user and assistant \cite{daniele2023amplify-instruct, bai2022traininghelpfulharmlessassistant}. 

\begin{figure}[t!]
    \centering
    \includegraphics[width=0.8\linewidth]{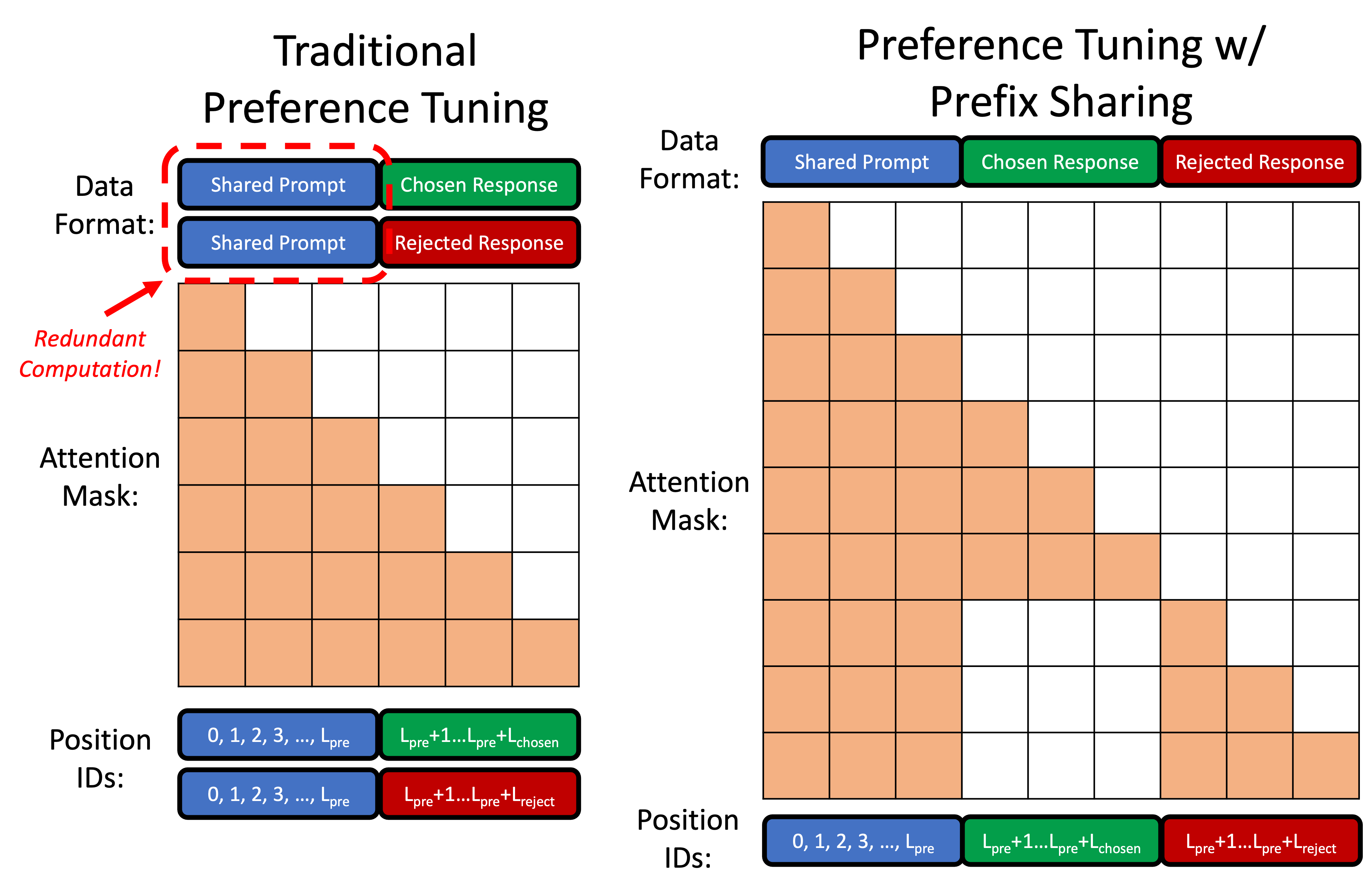}
    \caption{\textbf{Method overview.} Prefix sharing removes redundant computation of the shared prompt prefix by combining the responses into a single sequence and modifying the attention mask to prevent cross-response contamination. }
    \label{fig:matching}
\end{figure}

To eliminate this redundancy, we propose prefix sharing for preference tuning, where chosen and rejected responses are processed as one sequence with a shared prefix, as illustrated in Figure~\ref{fig:matching}. We employ a custom attention mask to delineate the two responses during the model's forward pass. Our approach accelerates training by significantly reducing total training tokens, performing best when training time scales roughly linearly with token count. For popular preference tuning datasets, we find that prefix sharing can enable roughly $1.1$-$1.5\times$ improvement to training throughput, without any effect on convergence and with a small reduction in memory consumption. We observe speedups of about $1.5\times$  for datasets that have long prefixes and high prefix-to-completion length ratios.

To further improve training throughput we implement sequence packing \cite{krell2021efficient} for prefix-shared inputs. Compared to sequence packing for the baseline paired input format, we find that with prefix sharing, packing can add additional efficiency gains, especially for datasets with smaller overall sequence lengths, enabling our method to provide more consistent $1.3$-$1.6\times$ speedups.

In our experiments, we focus mainly on Direct Preference Optimization (DPO) \cite{rafailov2024directpreferenceoptimizationlanguage} as it is currently the most popular offline paired preference optimization algorithm. However, our insights into prefix sharing for DPO can be easily transferred to other similar paired preference tuning methods.

\section{Background}

\subsection{Preference Optimization}

Preference optimization methods enable LLMs to be optimized for arbitrary goals defined by binary preference data, where pairs of responses to a shared prompt are ranked. Traditional on-policy RL-based approaches use preference data to train a reward model that assigns scalar scores to LLM responses \cite{ouyang2022traininglanguagemodelsfollow,stiennon2022learningsummarizehumanfeedback}. The LLM ($\pi_{\theta}$) is then trained using an online RL algorithm such as PPO~\cite{schulman2017proximalpolicyoptimizationalgorithms} to maximize the reward model while also being regularized to prevent significant deviation from the original reference LLM ($\pi_{\text{ref}}$).

Direct Preference Optimization (DPO) simplifies this multi-stage process by reparameterizing the reward model in terms of $\pi_{\theta}$ and $\pi_{\text{ref}}$ directly. For a single paired preference data sample, where $x$ is the prompt and $y_{\text{chosen}}$, $y_{\text{rejected}}$ are the chosen and rejected completions, the DPO loss function is:
\begin{equation}
    \mathcal{L}(x, y_{\text{chosen}},y_{\text{rejected}}; \pi_\theta, \pi_{\text{ref}}) = -\log \sigma\left(\beta \frac{\log \pi_{\theta} (y_{\text{chosen}} | x)}{\log \pi_{\text{ref}} (y_{\text{chosen}} | x)} - \beta \frac{\log \pi_{\theta} (y_{\text{rejected}} | x)}{\log \pi_{\text{ref}} (y_{\text{rejected}} | x)}\right)
\end{equation}

To compute this loss function, we need to calculate the total log probabilities of each of the completions $y_{\text{chosen}}$ and $y_{\text{rejected}}$. Typical implementations of DPO accomplish this by formatting each training sample as two full input sequences which are batched together: the shared prompt followed by the chosen response, and the shared prompt followed by the rejected response \cite{vonwerra2022trl, hu2024openrlhfeasytousescalablehighperformance,rafailov2024directpreferenceoptimizationlanguage,ivison2024unpacking}. The log probabilities are then calculated and extracted for each response independently, and so the computation for the shared prompt is repeated twice for every training sample. Since the shared prompt's activations are the exact same for both sequences due to causal masking, it is unnecessary to compute the shared prompt twice. In this work, we leverage prefix sharing to remove this redundancy while ensuring that the log probabilities are computed identically.

\subsection{Prefix Sharing for Inference}
While there has not been much exploration surrounding prefix sharing for training, there have been some works that explore prefix sharing for LLM decoding at inference time \cite{juravsky2024hydragenhighthroughputllminference,athiwaratkun2024bifurcatedattentionacceleratingmassively,cascade-inference}, where a prompt prefix is shared across multiple sequences that are being decoded. Since decoding is often memory-bound \cite{shazeer2019fasttransformerdecodingwritehead}, these works focus on reducing memory reads of the precomputed prefix KV-cache using custom attention kernel implementations.

On the other hand, LLM training at sufficient batch sizes is mostly compute-bound \cite{xia2024understandingperformanceestimatingcost}, and thus most of the gains from prefix sharing for DPO training are from directly reducing the number of total training tokens, rather than speedups from improving self-attention. Because of this, we use a simpler approach for the self-attention operation, employing a custom attention mask rather than designing a custom kernel, making our method very flexible and easily extensible to optimizations such as sequence packing. Our approach is most similar to the Medusa speculative decoding method~\cite{cai2024medusasimplellminference}, where a custom attention mask is used to simultaneously decode multiple sequences in a tree structure.

\section{Methodology}
\subsection{Prefix Sharing}
Instead of separating the paired responses into two different sequences, we place the prompt, chosen response, and rejected response into a single sequence, as shown in Figure~\ref{fig:matching}. In the attention mask, we mask out the region where the rejected response attends to the chosen response, ensuring that the log probabilities of both responses are computed independently of each other. Therefore, this attention computation is identical to the normal paired data format since the chosen and rejected responses can independently attend to the shared prompt. We also set the position IDs so that the sequences are the same as the normal paired format. 



We use PyTorch's FlexAttention~\cite{he2024flexattention} to implement the attention layers. Using FlexAttention's \texttt{mask\_mod}, we can leverage block sparsity to skip fully masked-out blocks of our custom attention mask. This enables us to skip computation of the masked region where the rejected response attends to the chosen response. At the beginning of every training step (i.e for every mini-batch), we compute the sparse block mask for our custom attention scheme and pass this block mask into each FlexAttention layer.

FlexAttention's \texttt{mask\_mod} constructs the custom attention mask using a user-defined function that outputs a boolean for a given batch sample index, query index, and key-value index. We implement this function by keeping track of the start indices of the chosen and rejected completions for each sample (see Algorithm~\ref{alg:dpomask}).

\algnewcommand{\LineComment}[1]{\State \(\triangleright\) \textit{#1}}
\begin{algorithm}
\footnotesize
\caption{Prefix Sharing Mask}\label{alg:dpomask}
\begin{algorithmic}
\item \textbf{Given: }sequence length $L$, batch size $B$
\item \textbf{Input: }batch sample index $b$; query index $q$; key-value index $kv$; per-token chosen and rejected completion start indices $I_{\text{chosen}},I_{\text{rejected}} \in \{0\ldots L-1\}^{B\times L}$
\item 
\LineComment{apply standard causal mask, masked regions are set to False}
\State $\textbf{causal\_mask} = (kv \leq q)$
\item
\LineComment{prevent queries in rejected from attending to keys in chosen}
\State $\text{chosen\_ind} = I_{\text{chosen}}[b]$
\State $\text{rejected\_ind} = I_{\text{rejected}}[b]$
\State $\textbf{dpo\_mask} = !((\text{rejected\_ind} \leq q) \text{ \&\& }(\text{chosen\_ind} \leq kv < \text{rejected\_ind}))$ 
\item
\item \Return $\textbf{causal\_mask}\text{ \&\& }\textbf{dpo\_mask}$
\end{algorithmic}
\end{algorithm}


\subsection{Sequence Packing}
Sequence packing is a popular technique for improving training efficiency in NLP \cite{liu2019roberta, krell2021efficient}. Typically, an approximate solution to the bin-packing algorithm is used to sample sequences to pack in a mini-batch. While sequence packing for the regular causal language modeling task is well studied, packing for paired-preference data is more involved. Since we have two sequences (chosen and rejected) per training sample, each with varying length, one cannot pack these independently. Thus, a simple strategy could be to just combine the chosen and rejected sequences (i.e. the shared prompt along with the response) for a sample and to treat this as a single unit for packing. As shown in Algorithm~\ref{alg:dpobaselinepackingmask}, one would need to also keep track of the boundaries of the different responses in the batch to prevent cross-contamination. The chosen and rejected responses for each sample are also recovered at loss calculation with these boundaries. 

\begin{figure}
    \centering
    \includegraphics[width=0.8\linewidth]{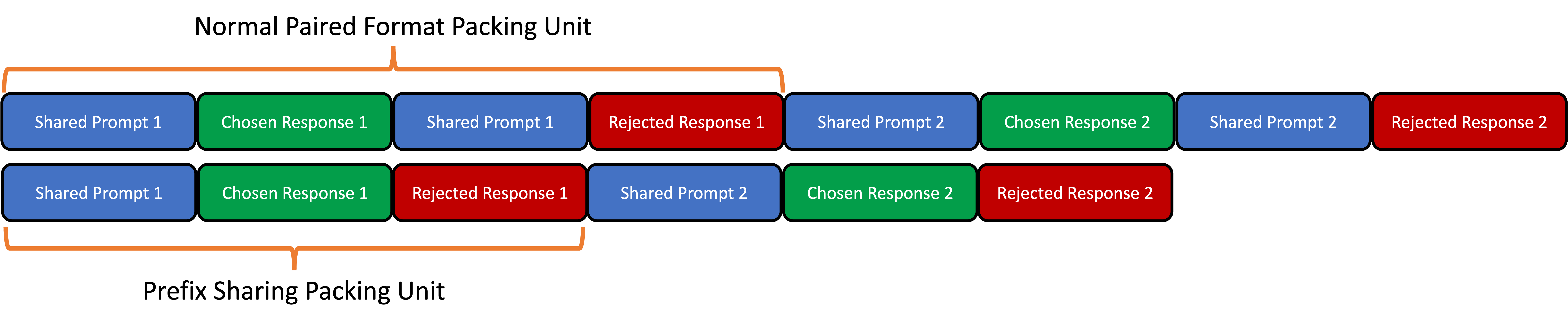}
    \caption{Sequence packing with and without prefix-sharing for paired preference inputs, illustrated for two training samples. Without prefix-sharing, a sequence packing implementation will have to treat the chosen and rejected responses, each prefixed by the common prompt, as a single unit and then pack these units together. With prefix sharing, the unit for sequence packing is now the shared prompt with the chosen and rejected response.}
    \label{fig:sequence-packing}
\end{figure}

Sequence packing for our prefix sharing method is similarly straightforward (see Figure~\ref{fig:sequence-packing}). Without a redundant shared prefix, the maximum sequence length for each packing unit is much shorter. In Section~\ref{sec:packingbenchmarks}, we show that packing with prefix sharing can lead to much better efficiency gains than prefix sharing alone. We further demonstrate that prefix sharing with packing has no negative impact on model quality in Appendix~\ref{sec:packing-training}.

To implement sequence packing for prefix sharing without cross-sequence contamination, we again leverage FlexAttention and further customize the block mask. As shown in Algorithm~\ref{alg:dpopackingmask}, we use a \texttt{mask\_mod} function that is similar to non-packing prefix sharing, except we also keep track of unique document IDs for each sample and store the chosen/rejected start indices as a dense $B\times L$ array.

\begin{algorithm}
\footnotesize
\caption{Attention Mask for Packing w/o Prefix Sharing}\label{alg:dpobaselinepackingmask}
\begin{algorithmic}
\item \textbf{Given: }sequence length $L$, batch size $B$
\item \textbf{Input: }batch sample index $b$; query index $q$; key-value index $kv$; per-token response IDs $R \in \mathbb{Z}^{B\times L}$
\item 
\LineComment{apply standard causal mask, masked regions are set to False}
\State $\textbf{causal\_mask} = (kv \leq q)$
\item
\LineComment{apply masking to prevent tokens from separate samples and different responses from attending to each other}
\State $\textbf{response\_mask} = (R[b][q] == R[b][kv])$
\item
\item \Return $\textbf{causal\_mask}\text{ \&\& }\textbf{response\_mask}$
\end{algorithmic}
\end{algorithm}

\begin{algorithm}
\footnotesize
\caption{Prefix Sharing Mask with Packing}\label{alg:dpopackingmask}
\begin{algorithmic}
\item \textbf{Given: }sequence length $L$, batch size $B$
\item \textbf{Input: }batch sample index $b$; query index $q$; key-value index $kv$; per-token chosen and rejected completion start indices $I_{\text{chosen}},I_{\text{rejected}} \in \{0\ldots L-1\}^{B\times L}$; document IDs $D \in 	\mathbb{Z}^{B\times L}$
\item 
\LineComment{apply standard causal mask, masked regions are set to False}
\State $\textbf{causal\_mask} = (kv \leq q)$
\item
\LineComment{apply document masking to prevent tokens in separate samples from attending to each other}
\State $\textbf{document\_mask} = (D[b][q] == D[b][kv])$
\item
\LineComment{prevent queries in rejected from attending to keys in chosen}
\State $\text{chosen\_ind} = I_{\text{chosen}}[b][q]$
\State $\text{rejected\_ind} = I_{\text{rejected}}[b][q]$
\State $\textbf{dpo\_mask} = !((\text{rejected\_ind} \leq q) \text{ \&\& }(\text{chosen\_ind} \leq kv < \text{rejected\_ind}))$
\item
\item \Return $\textbf{causal\_mask}\text{ \&\& }\textbf{document\_mask}\text{ \&\& }\textbf{dpo\_mask}$
\end{algorithmic}
\end{algorithm}

\section{Experiments}\label{sec:exps}

\subsection{Baselines}

We compare our prefix sharing approach against two baselines: FlexAttention with the normal paired data format and FlashAttention-3~\cite{shah2024flashattention3fastaccurateattention} with the normal paired data format. Despite FlexAttention's advantages, its base performance is worse compared to less general attention kernel implementations such as FlashAttention-3, which does not easily support arbitrary attention masks (and thus can not be used with prefix sharing) but is more optimized than FlexAttention. Since the normal paired format is compatible with FlashAttention-3 (as it only requires causal masks), while prefix sharing is only compatible with FlexAttention, we include FlashAttention-3 with the normal paired format as a baseline in our benchmarking experiments.

\subsection{Individual Layer Micro-benchmarking}

We perform microbenchmarks in three settings: the MLP layer, the self-attention computation, and the full Attention layer (including linear projections) of Mistral-7B~\cite{jiang2023mistral}. For each experiment we report the sum of the forward and backward pass time, sweeping across different prefix and completion lengths. All of the following microbenchmarking experiments were conducted on a single NVIDIA H100 GPU.

\subsubsection{MLP Micro-benchmarking}
\begin{figure}[h!]
    \centering
    \includegraphics[width=0.49\linewidth]{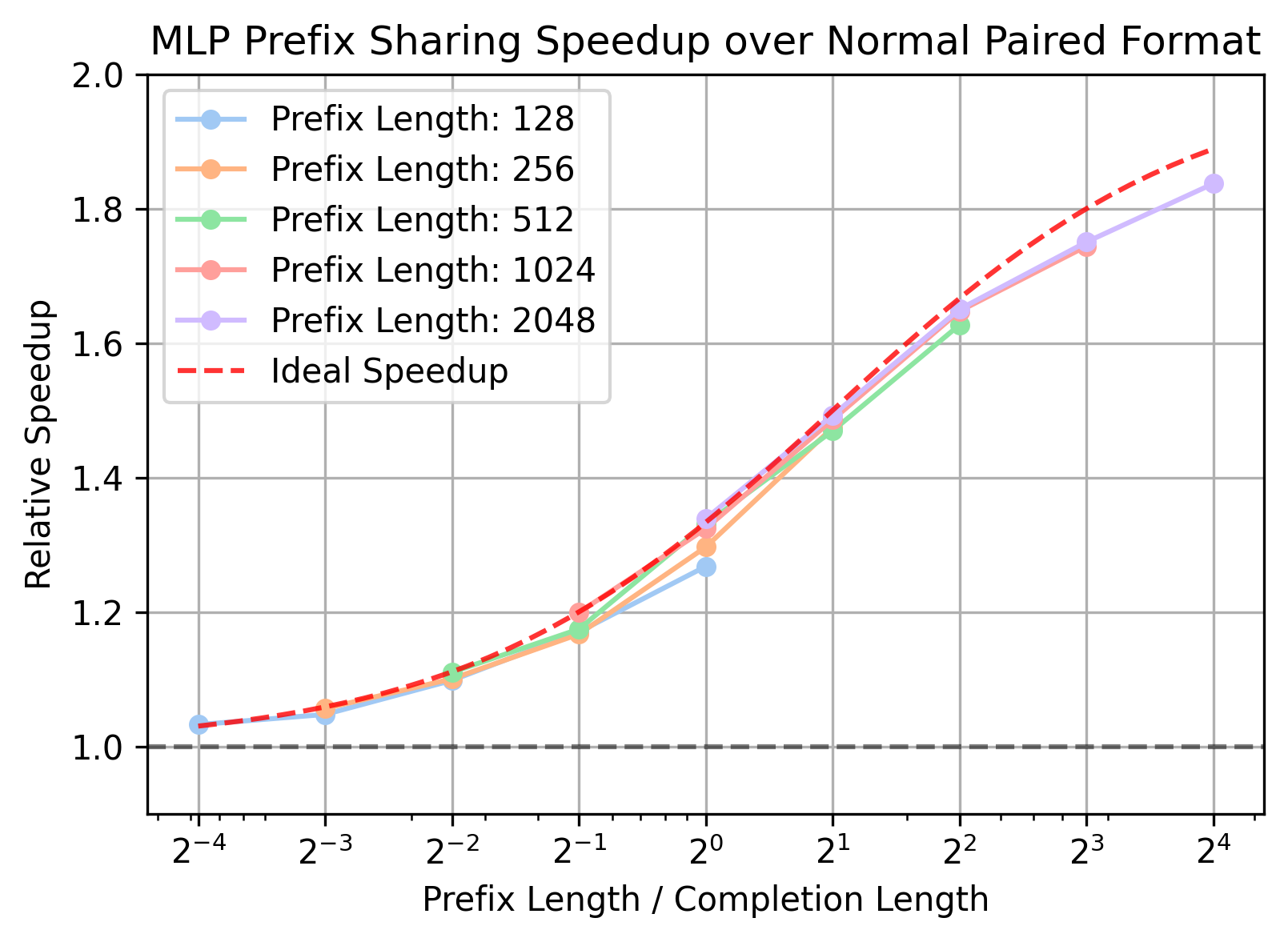}
    \caption{Microbenchmarking results of the MLP layer for Mistral 7B. Relative speedups of prefix sharing over normal paired data are shown in comparison to the ideal speedup (assuming linear runtime). We see that the MLP layer scales very closely to the ideal speedup and that increasing the prefix length helps push the speedup closer to the ideal for a given prefix to completion ratio.}
    \label{fig:h100mlpmicro}
\end{figure}

For the MLP layer experiments, we compare the prefix sharing data format to the normal paired format. As shown in Figure~\ref{fig:h100mlpmicro}, we find that for longer prefix lengths, we get nearly ideal linear speedups\footnote{We compute the ideal linear speedup for a given prefix length $p$ and completion length $c$ as $\frac{2(p + c)}{p + 2c}$.} from the reduction in total tokens due to the operation being mostly compute-bound. For the shorter prefix lengths of 128 and 256, the improvements are still valuable but are smaller due to constant time overheads (such as memory reads of the weights) taking up a larger proportion of the total time.

\subsubsection{Attention Micro-benchmarking}

For the attention experiments, we compare FlexAttention with the prefix sharing format to both FlexAttention and FlashAttention-3 with the normal paired format.

\begin{figure}[h!]
    \centering
    \includegraphics[width=0.49\linewidth]{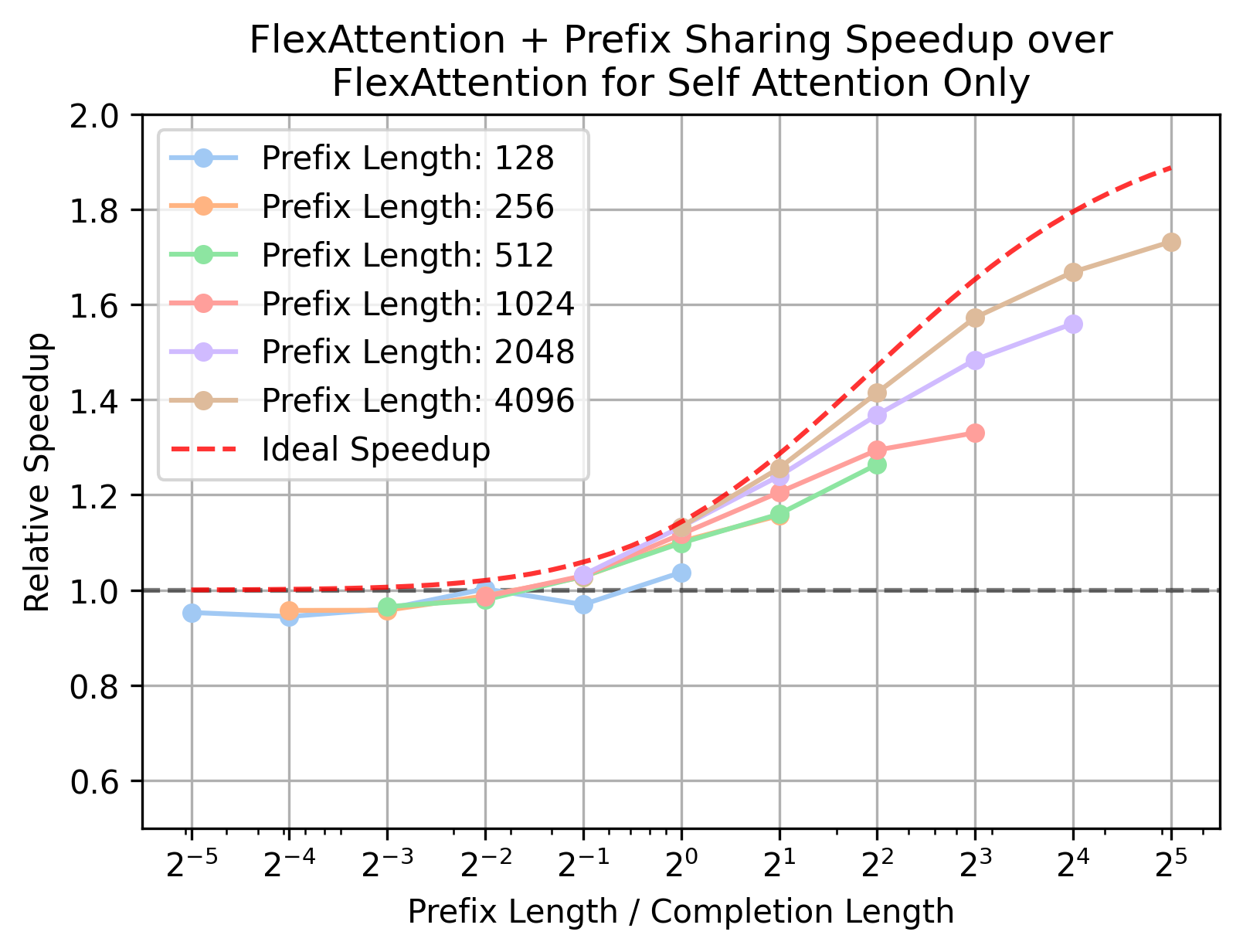}
    \includegraphics[width=0.49\linewidth]{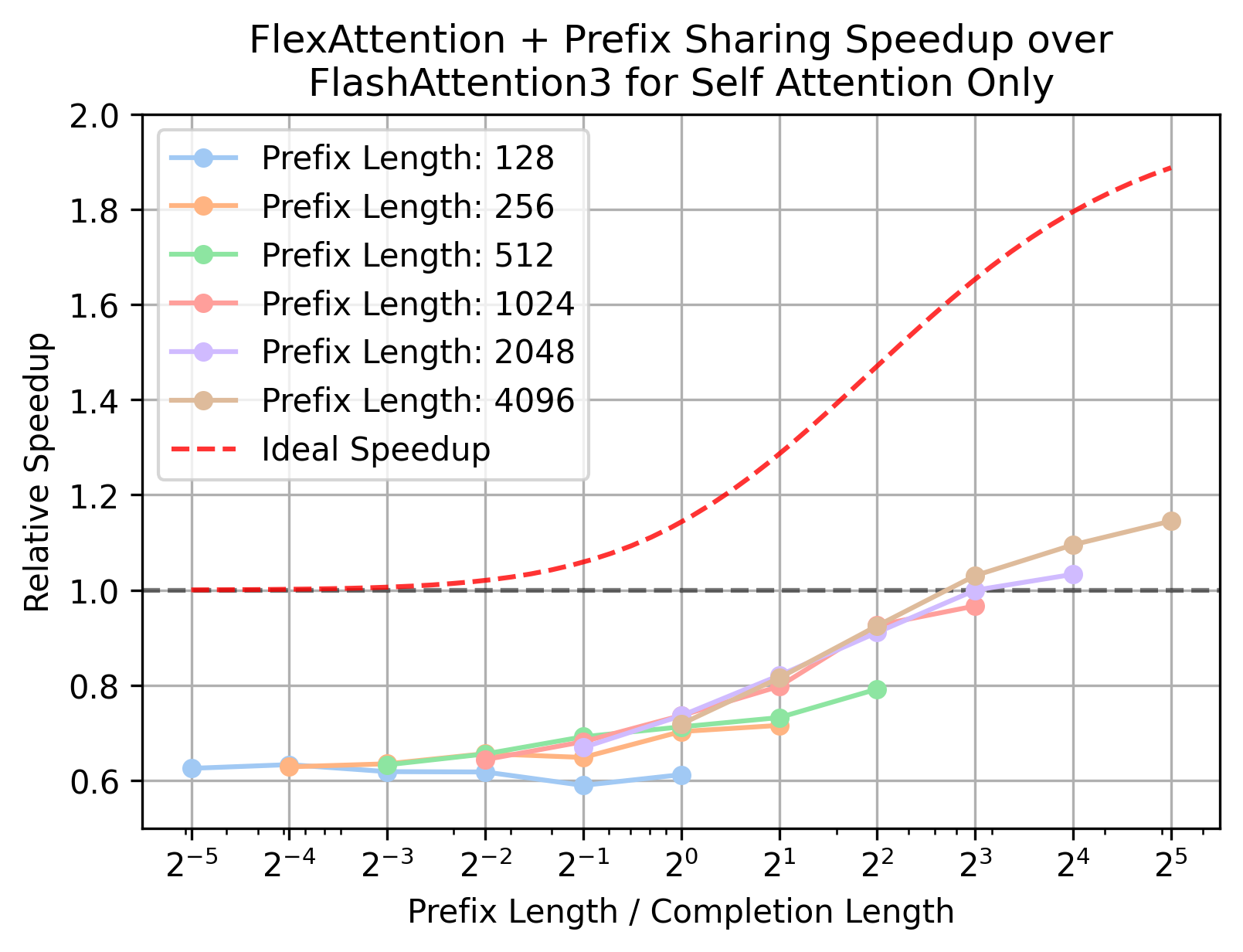}
    \caption{Microbenchmarking results of the self-attention operation only for Mistral 7B. Relative speedups of FlexAttention with prefix sharing over FlashAttention-3 and FlexAttention are shown, along with the ideal speedup (assuming perfect quadratic scaling). We see that for high prefix lengths, FlexAttention with prefix sharing attains nearly ideal speedups over FlexAttention without prefix sharing, but overall it is still slower or similar in speed to FlashAttention-3. Nevertheless, we find in practice that self-attention contributes little to overall training time and thus has minimal impacts.}

    \label{fig:h100selfattnmicro}
\end{figure}

In Figure~\ref{fig:h100selfattnmicro}, we can see that FlexAttention with prefix sharing shows clear trends towards higher performance boosts at greater prefix to completion length ratios, attaining close to ideal speedups\footnote{We compute the ideal quadratic attention speedup for prefix length $p$ and completion length $c$ as $\frac{2(p + c)^2}{2(p + c)^2 - p^2}$.} over FlexAttention with normal paired data. However, when compared to FlashAttention-3, FlexAttention tends to be much slower, only matching or exceeding performance at ratios that are $\geq$8.

\begin{figure}[h!]
    \centering
    \includegraphics[width=0.49\linewidth]{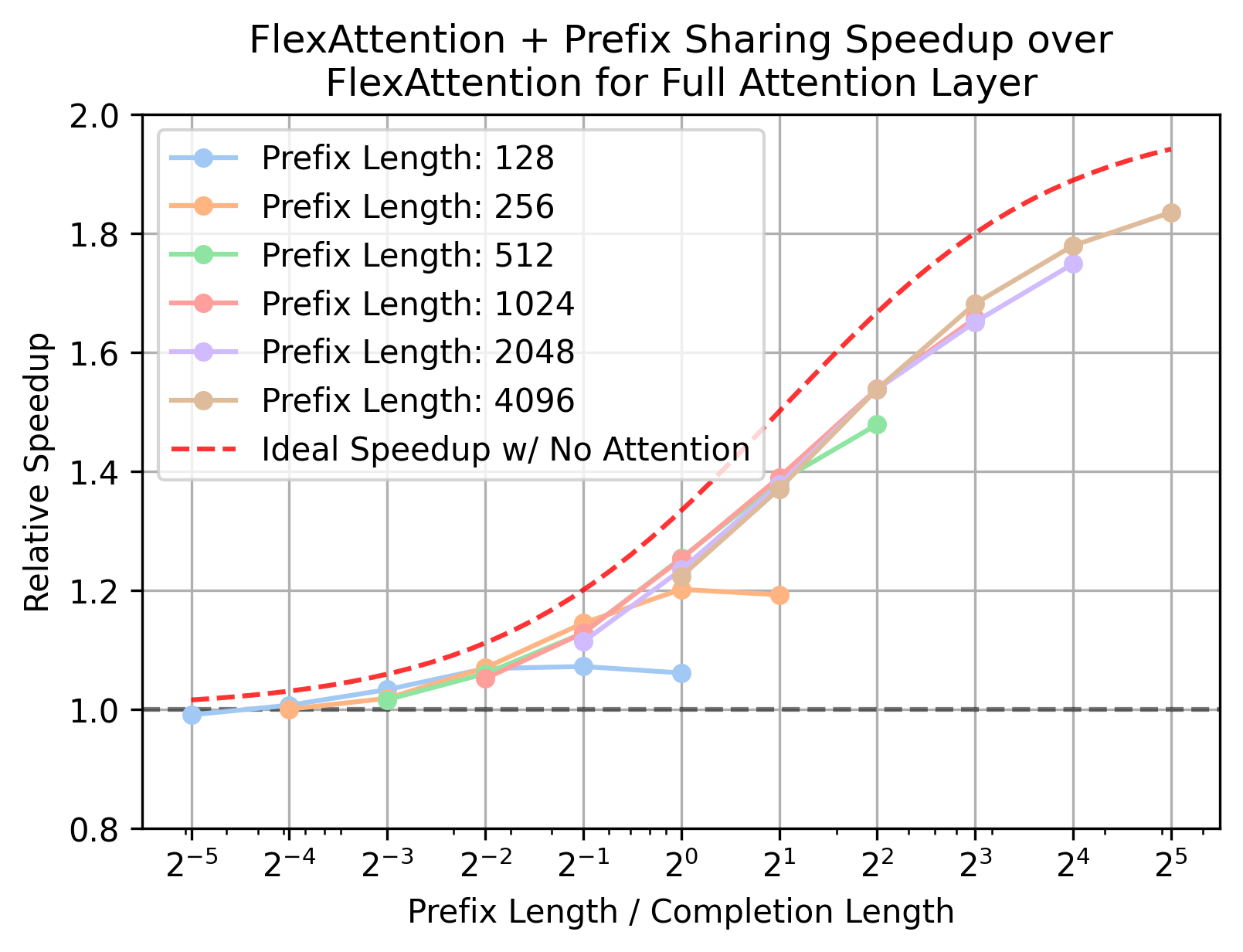}
    \includegraphics[width=0.49\linewidth]{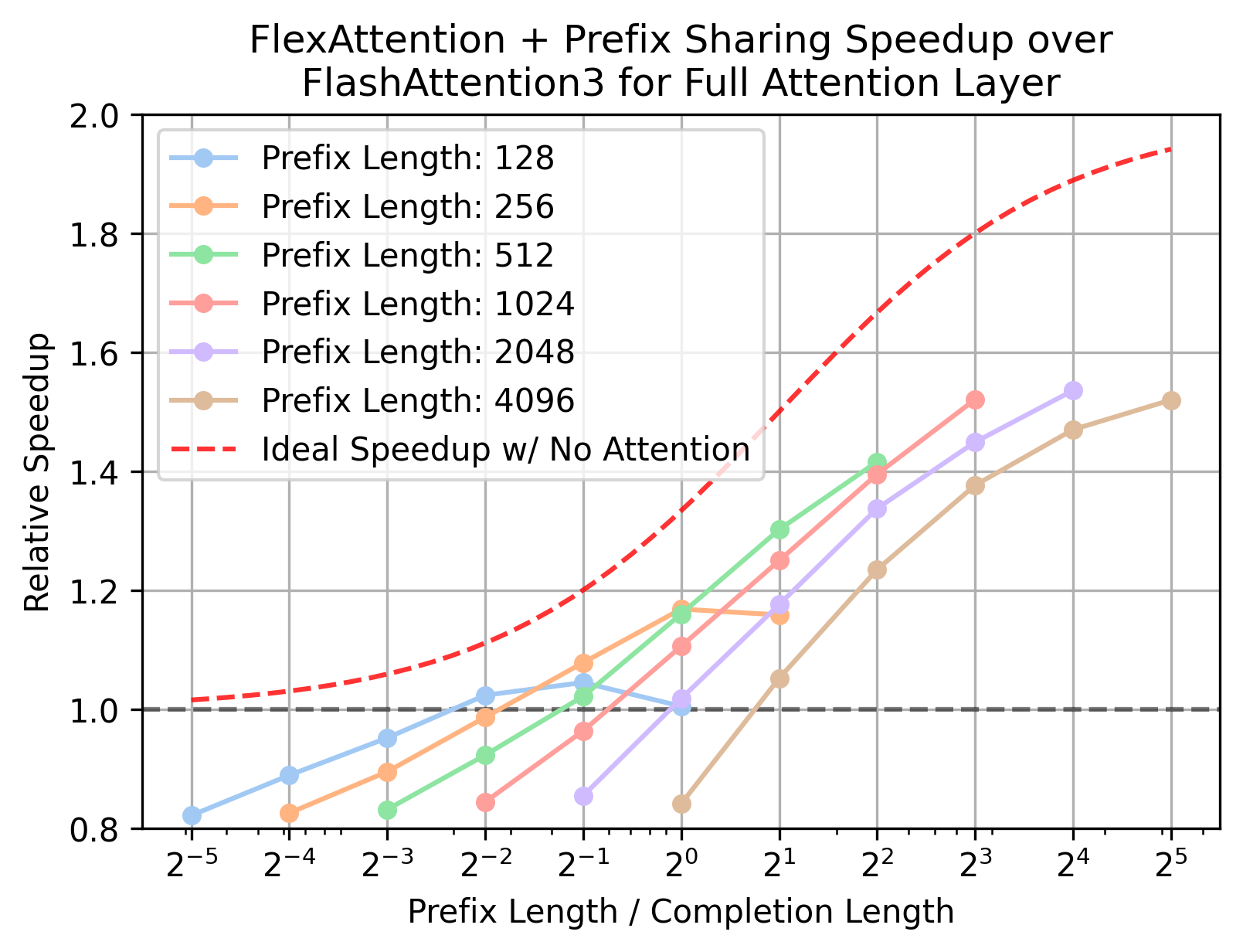}
    \caption{Microbenchmarking results of the full self-attention layer (QKV projection + self-attention) for Mistral 7B. Relative speedups of FlexAttention with prefix sharing over FlashAttention-3 and FlexAttention are shown, along with the ideal speedup (assuming linear runtime). We see that although FlexAttention is slower than FlashAttention-3 for lower ratios between the prefix and completion length, as the ratio grows, FlexAttention with prefix sharing become faster.}
    \label{fig:h100attnmicro}
\end{figure}

Nevertheless, in Figure~\ref{fig:h100attnmicro} we illustrate that when benchmarking the full attention layer (QKV projection + self-attention), the discrepancy between FlexAttention and FlashAttention-3 is diluted significantly since the self-attention operation is a small proportion of the overall speed. For the full attention layer, FlexAttention with prefix sharing is faster than FlashAttention-3 for most sequences with ratios $\geq$1, thus making prefix sharing valuable for many datasets despite the worse base performance of FlexAttention.

Overall, these results suggest that our approach is most promising for longer sequences where performance is mostly compute-bound, and for higher prefix length to response ratios where the total token reduction is greatest. Although FlashAttention-3 is inherently faster than FlexAttention, the effects are minimal as self-attention is a small proportion of the overall time, and in the following full model training experiments in Section~\ref{sec:fulltraining} we further validate this.

\subsection{Full Model Training Benchmarking}\label{sec:fulltraining}
To highlight the usefulness of our approach in real-world settings, we evaluate our approach on popular preference tuning datasets. We include five datasets with high prefix to completion ratios spanning from single-turn chat (Tulu-Helpsteer~\cite{ivison2024unpacking,wang2023helpsteermultiattributehelpfulnessdataset}) and multi-turn chat (Capybara~\cite{daniele2023amplify-instruct}, HH-RLHF~\cite{bai2022traininghelpfulharmlessassistant}) to specific domains like math~(MetaMath-DPO~\cite{pal2024smaugfixingfailuremodes,yu2024metamathbootstrapmathematicalquestions}) and summarization (TLDR \cite{stiennon2022learningsummarizehumanfeedback}). We also include Ultrafeedback \cite{cui2024ultrafeedbackboostinglanguagemodels}, a single-turn chat dataset with low prefix to completion ratios, to showcase how our approach still leads to small speedups for low ratios.

We base our DPO training implementation on the open-source TRL library~\cite{vonwerra2022trl} and use DeepSpeed ZeRO-3~\cite{rasley2020deepspeed} to accelerate training. All our experiments are conducted on 1 8xH100 instance with CUDA 12.3 and PyTorch 2.5.0.
\subsubsection{Prefix Sharing Benchmarking}\label{sec:prefixsharingfulltraining}
We use a per-device batch size of 4 and fix the maximum prompt and total sequence length based on dataset statistics to minimize thresholding. For the FlexAttention experiments, we use our own attention layer implementation that leverages FlexAttention and constructs a block-sparse attention mask at the beginning of each training step.


	

%
%
%
%
%
%

\begin{table}[h!]
\centering
\small
\begin{tabular}{lcc|cc|c}
\toprule
Dataset  & \shortstack{Median\\Overall Len} & \shortstack{Prefix /\\ Completion}& FA3 & \shortstack{Flex\\Attn} & \shortstack{Flex +\\Prefix Sharing} \\
\midrule
Capybara~\cite{daniele2023amplify-instruct} & $1160$ & $1.59$ &  $8.38$ & $7.75$ & $11.90$\hspace{1.5mm}($1.42\times$, $1.54\times$) \\
HH-RLHF~\cite{bai2022traininghelpfulharmlessassistant} & $186$ & $2.15$ & $33.71$ & $30.25$ & $36.11$\hspace{1.5mm}($1.07\times$, $1.19\times$) \\
MetaMath-DPO~\cite{pal2024smaugfixingfailuremodes,yu2024metamathbootstrapmathematicalquestions}  & $872$ & $3.91$ & $13.86$ & $13.02$ & $19.13$\hspace{1.5mm}($1.38\times$, $1.47\times$) \\
TLDR~\cite{stiennon2022learningsummarizehumanfeedback}  & $416$ & $11.14$ & $31.43$ & $29.53$ & $35.36$\hspace{1.5mm}($1.12\times$, $1.20\times$) \\
Tulu-Helpsteer~\cite{ivison2024unpacking,wang2023helpsteermultiattributehelpfulnessdataset} & $775$ & $6.34$ & $14.83$ & $13.93$ & $21.75$\hspace{1.5mm}($1.47\times$, $1.56\times$)  \\
Ultrafeedback \cite{cui2024ultrafeedbackboostinglanguagemodels}  & $409$ & $0.42$ & $18.40$ & $17.31$ & $20.46$\hspace{1.5mm}($1.11\times$, $1.18\times$) \\
\bottomrule
\end{tabular}
\vspace{2mm}
\caption{Comparison of training samples per second for different attention implementations. Relative speedups over FlashAttention-3 and FlexAttention, respectively, are shown for the FlexAttention with Prefix Sharing column. FlexAttention with prefix sharing consistently outperforms the baselines, with speedups ranging from $1.1$-$1.5\times$, with FlexAttention alone being slower than FA3. For the Prefix / Completion column, we report the median ratio. For high median overall lengths ($> 500$), the gains from prefix sharing are $>35\%$, with better gains for high prefix to completion ratios.}
\label{tab:training_times}
\end{table}

Training throughput in samples per second\footnote{We report samples per second instead of tokens per second, since "tokens processed" can be confusing across methods} is shown in Table~\ref{tab:training_times}. We find that FlexAttention with prefix sharing consistently speeds up training time, even though FlexAttention alone is generally slower than FlashAttention-3. We further observe a small reduction in memory consumption with the paired data format, as is expected given the overall reduction in input (and thus activation) size. Our approach is especially effective for sequences that have high ratios and long overall lengths, enabling us to reach gains of up to $1.4$-$1.5\times$ over FlashAttention-3. Even for Ultrafeedback, which has a very low ratio, we see some speedups, showing that our method is helpful even for low prefix to completion ratio settings.

However, we can see that for datasets like HH-RLHF and TLDR which have high ratios but low overall lengths, our speedups are smaller. To address this, we show in the following Section~\ref{sec:packingbenchmarks} that sequence packing helps remedy this, pushing training closer towards linear compute bound scaling by increasing the effective overall sequence length. 
\subsubsection{Packing Benchmarking}\label{sec:packingbenchmarks}

We use a First-Fit-Decreasing (FFD) based sampler for efficient sequence packing\footnote{\texttt{https://github.com/imoneoi/multipack\_sampler}}. For both the FlexAttention-based baseline and our prefix sharing algorithm, we add a document ID-based attention mask to prevent cross-contamination. For FlashAttention-3, we use the packed position IDs (which is internally translated to an array of cumulative sequence lengths) to track the boundaries of different responses (and different documents) in a packed mini-batch. We set the packing length to be the desired batch size, $\texttt{bsz}$ (set to be the same as before), multiplied by the maximum sequence length~($\texttt{seq\_len})$ of the packing unit in the dataset. This provides higher packing efficiency than packing until $\texttt{seq\_len}$ and sampling $\texttt{bsz}$ number of packed sequences. We pad all batches to the fixed packing length for better performance.

We show our results in Table $\ref{tab:training_times_packing}$. Overall, packing with prefix sharing consistently outperforms the baselines, with most datasets achieving about $1.3$-$1.6\times$ improvement in training throughput compared to FlashAttention-3 with packing. Packing provides a significant boost in training throughput for datasets with low median overall lengths (such as HH-RLHF, TLDR) compared to the non-packing results. Since prefix sharing decreases the number of tokens in each packing unit (see Figure~\ref{fig:sequence-packing}), sequence packing is more effective here than the baselines. When comparing the speedups with and without packing, we observe that the relative improvement over the FlashAttention-3 baseline increases from $1.07\times$ to $1.41\times$ for HH-RLHF (a $32\%$ increase), and from $1.12\times$ to $1.35\times$ for TLDR (a $21\%$ increase). UltraFeedback also still sees performance boosts ($1.11\times$ to $1.17\times$) with packing despite having a low prefix-to-completion ratio.

\begin{table}[]
\small
    \centering
    \begin{tabular}{lcc|cc|c}
    \toprule
\shortstack{Dataset\\Name} & \shortstack{Median\\Overall Len} & \shortstack{Prefix /\\ Completion} & \shortstack{FA3 +\\Packing} & \shortstack{Flex Attn +\\Packing} & \shortstack{Flex Attn +\\Prefix Sharing +\\Packing} \\
\midrule
    Capybara~\cite{daniele2023amplify-instruct} & $1160$ & $1.59$ &  $17.89$ & $17.63$ & $23.89$\hspace{1.5mm}($1.34\times$, $1.36\times$) \\
    HH-RLHF~\cite{bai2022traininghelpfulharmlessassistant} & $186$ & $2.15$ &  $109.77$ & $104.99$ & $155.04$\hspace{1.5mm}($1.41\times$, $1.48\times$) \\
    MetaMath-DPO~\cite{pal2024smaugfixingfailuremodes,yu2024metamathbootstrapmathematicalquestions}&  $872$ & $3.91$ & $24.21$ & $23.83$ & $38.07$\hspace{1.5mm}($1.57\times$, $1.60\times$) \\
    TLDR~\cite{stiennon2022learningsummarizehumanfeedback} &  $416$ & $11.14$ & $44.11$ & $43.22$ & $59.76$\hspace{1.5mm}($1.35\times$, $1.38\times$) \\
    Tulu-Helpsteer~\cite{ivison2024unpacking,wang2023helpsteermultiattributehelpfulnessdataset} &  $775$ & $6.34$ & $29.85$ & $28.98$ & $44.10$ ($1.48\times$, $1.52\times$)\hspace{1.5mm}\\
    Ultrafeedback~\cite{cui2024ultrafeedbackboostinglanguagemodels}  &  $409$ & $0.42$ & $45.46$ & $44.13$ & $53.21$\hspace{1.5mm}($1.17\times$, $1.21\times$) \\
    \bottomrule
    \end{tabular}
    \vspace{2mm}
    \caption{Comparison of training samples per second with sequence packing. For Flex Attn + Prefix Sharing + Packing, relative speedups over FA3 + Packing and Flex Attn + Packing are shown in parentheses, respectively. For the Prefix / Completion column, we report the median ratio. Our method (Prefix sharing + Packing) demonstrates at least a $30\%$ increase in training throughput for most datasets. The impact of sequence packing is especially prominent for datasets like HH-RLHF and TLDR with shorter overall sequence lengths. Only Ultrafeedback, which has a extremely low prefix-to-completion ratio ($0.3$), shows a modest improvement of $21\%$ over the FlexAttention baseline.}
    \label{tab:training_times_packing}
\end{table}

\section{Conclusion}
In this work, we present prefix sharing, a simple but effective technique for improving training throughput in paired preference optimization. By processing chosen and rejected responses as one sequence with a shared prefix, our method achieves up to $1.5\times$ speedups on Direct Preference Optimization (DPO) datasets with high overall lengths and high prefix to completion length ratios. We further demonstrate the advantages of sequence packing with prefix sharing, enabling our method to benefit even datasets with smaller sequence lengths. While our experiments focus on DPO, our approach is applicable to other paired preference tuning methods, which we wish to explore in future work. 

\section*{Acknowledgements}

This research was supported with compute resources from Anyscale. We thank Will Lin and Sarah Schwettmann for their helpful feedback throughout this project.

\newpage
\bibliographystyle{plain} 
\bibliography{ref} 







\clearpage
\appendix
\section*{Supplementary Material}




\section{Packing Full Training Results}\label{sec:packing-training}

Naively applying packing with the same settings as non-packing could potentially lead to worse downstream performance, since for the same batch size the model trained with packing takes less gradient steps. Further, performance can be affected by cross-contamination across packed sequences. To verify that packing can match the performance of non-packing DPO, we evaluate the downstream performance of models trained with and without packing. To account for differences in effective batch size due to packing, we also sweep across different batch sizes.

We use the UltraFeedback dataset \cite{cui2024ultrafeedbackboostinglanguagemodels}, which is a large synthetically generated preference dataset that has been a popular choice for aligning open-source models such as Zephyr-7B-$\beta$ \cite{tunstall2023zephyrdirectdistillationlm}. Using the same hyperparameters as Zephyr, we finetune the Zephyr-7B-SFT model\footnote{The Zephyr model is trained by first applying SFT on Mistral-7B-v0.1 using the Ultrachat \cite{ding2023enhancingchatlanguagemodels} dataset, and then applying DPO with UltraFeedback.} on UltraFeedback using DPO, sweeping across different packing and batch size settings. Finally, we evaluate each model using MT-Bench \cite{zheng2023judgingllmasajudgemtbenchchatbot}, a popular multi-turn benchmark that judges conversational and instruction following abilities.

As shown in Table~\ref{table:mt_bench_scores}, we find that packing roughly matches or exceeds the normal paired format for MT-Bench scores across all batch sizes, thus showing that packing for DPO does not harm performance.


\begin{table}[ht]
\centering
\renewcommand{\arraystretch}{1.2}
\begin{tabular}{|l|c|c|}
\hline
\textbf{Method} & \textbf{Batch Size} & \textbf{MT-Bench Score} \\
\hline

\multirow{4}{*}{Packing + Prefix Sharing} & 32 & 7.3 \\
& 64 & \textbf{7.4} \\
& 96 & 7.3 \\
& 128 & 7.0 \\
\hline
\multirow{4}{*}{Normal Paired Format} & 32 & 7.0 \\
& 64 & 7.1 \\
& 96 & 7.2 \\
 & 128 & 7.1 \\
\hline
Baseline (Zephyr-7B-SFT) & N/A & 6.4 \\
\hline
\end{tabular}
\vspace{4pt}
\caption{MT-Bench \cite{zheng2023judgingllmasajudgemtbenchchatbot} scores for different packing and non-packing DPO training across different batch sizes. Models were trained with Ultrafeedback using hyperparameters from Zephyr \cite{tunstall2023zephyrdirectdistillationlm}.}
\label{table:mt_bench_scores}
\end{table}

\end{document}